\documentclass[journal]{IEEEtran}
\usepackage{epsfig,subfigure, float}

\begin{document}

\title{New Principles of Coordination in Large-scale \\Micro- and Molecular-Robotic Groups}

\author{S.~Kornienko, O.~Kornienko \\
{Institute of Parallel and Distributed Systems, University of Stuttgart,} \\
{Universit{\"a}tsstr.~38, D-70569 Stuttgart, Germany}
\thanks{IARP - IEEE/RAS - EURON Joint Workshop on MICRO \& NANO ROBOTICS, Paris, 23 - 24 October, 2006. This is a post-proceeding version of this work.}
}
\date{}
\maketitle

\begin{abstract}
Micro- and molecular-robotic systems act as large-scale swarms. Capabilities of sensing, communication and information processing are very limited on these scales. This short position paper describes a swarm-based minimalistic approach, which can be applied for coordinating collective behavior in such systems.
\end{abstract}

\section{Introduction}

Micro-robots and currently molecular-robotics~\cite{Balzani03} become important and extremely challenging branch of modern robotic research. The domain of micro- and molecular-robotic applications, such as biotechnologies, micro-systems construction, molecular engineering, and finally nanotechnologies, represents a huge economic, social and technological potential with a great impact on everyday human life~\cite{GOLEM}. Due to a large number of micro-objects, usually they are of hundred thousands, there is a need of such robotic systems, which can provide an autonomous and half-autonomous handling of many small objects. The main focus lies on a high-parallelization of handling processes performed by collectively working robots and involves different plan-generating~\cite{Kornienko_S04} and planning \cite{Kornienko_S03A} approaches.

Micro-robots and especially molecular systems, due to a small size, are very limited in actuation, sensing and communication~\cite{Kornienko_S05e}. However working in a collective way requires coordination of robotic behavior. The technological, size-dependent and functional restrictions hinder applications of approaches known in a larger mini- and "normal-size" robotics. As demonstrated by a few current micro-robotic projects (robot size $2\times2\times2mm$) ~\cite{I-Swarm}, we need to find a new principles of coordination even in the micro-robotic group  consisting of 1000 robots.

An approach, which can be successfully applied to limited robotic systems, is based on the low-level properties of optical perception~\cite{Kornienko_S05d}, in particular IR light~\cite{KornienkoS05d}, as well as on involvement of such processes into e.g. collective decision making \cite{Kornienko_OS01} or cognitive processes~\cite{Kornienko_S05a}. More generally, different nonlinear processes and models~\cite{Levi99} can be used for such purposes. Further in Sec.~\ref{sec:motivation} we provide an example of typical problems in limited micro-systems such as collective information processing and in Sec.~\ref{sec:mechanisms} briefly discuss a mechanism for solution of this problem.

\section{Challenge of Collective Information Processing}
\label{sec:motivation}

The motivating experiment originated from the swarm research around the "Jasmine" robot~\cite{Kornienko_S05b}. We use this platform for testing different bio-inspired approaches, developing algorithms of controllable-emergent \cite{Kornienko_S04a} collective behavior based on limited sensing and communication capabilities. This robot measures $26\times26\times20mm$ ($30\times30\times20mm$ in the latest version), with the number of robots between 50 and 130 in different experiments~\cite{swarmrobot}.

This point of this experiment is related to information transfer in robotic groups~\cite{Kornienko_S04b}. Information about state of environment, coordination and decision making processes are transferred via RF communication globally in the field or service type of robotics. Such a circulation of information is necessary for keeping all robots aware about environment and about individual robot's intensions. This represents a basis for global coordination mechanisms. Mini- and micro- robotics uses primarily only local communication. These are e.g. optical, electromagnetic or even chemical mechanisms. The global information transfer, and thus coordination mechanisms, can be produced in this case only by a mechanism that propagates messages through multiple robot-robot connections. Parameters of a global circulation of information (such as a global propagation speed or global propagation time) depend on characteristics of local communication.
\begin{figure}[htp]
\centering 
\includegraphics[width=0.5\textwidth]{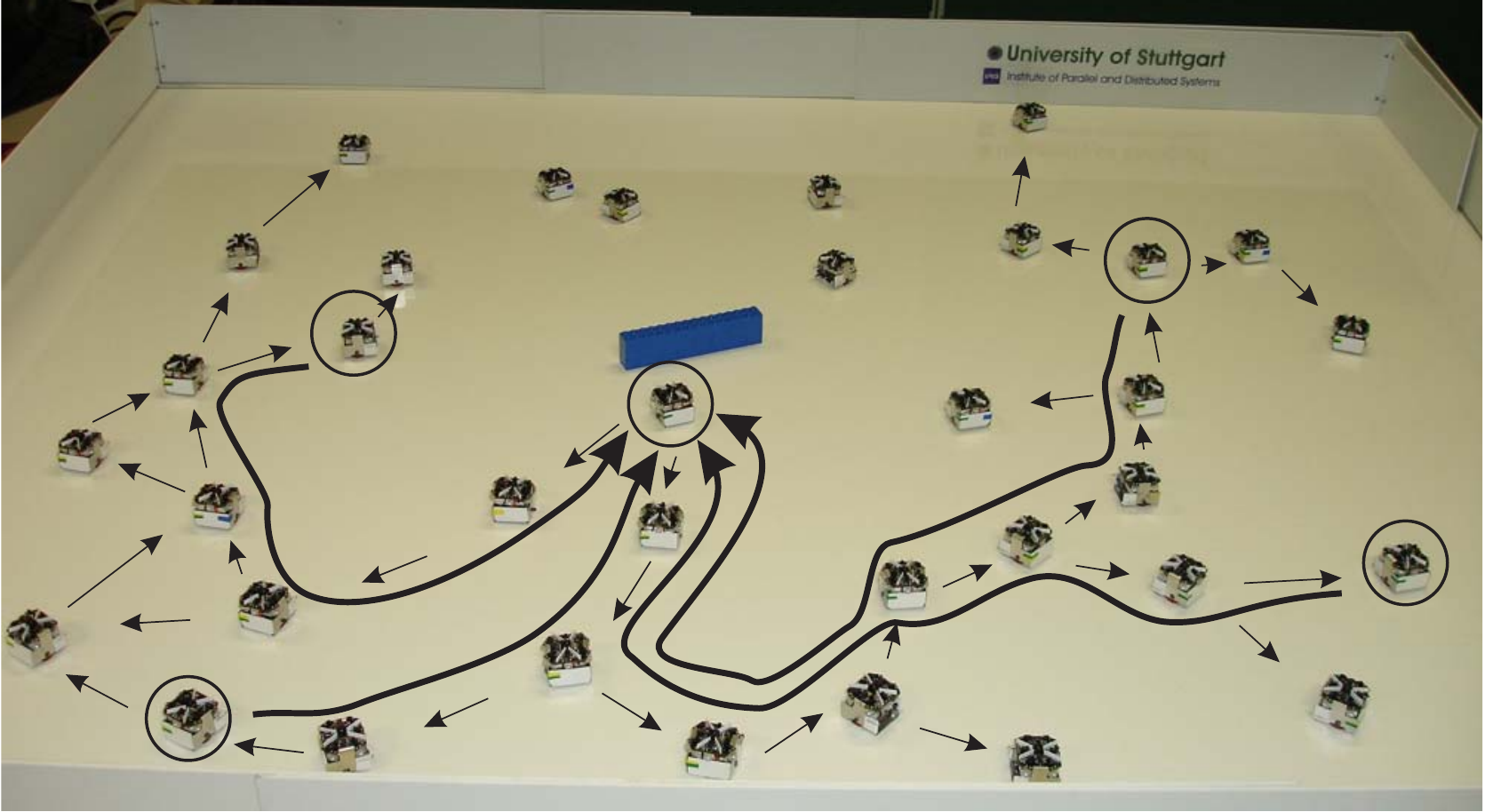}
\caption{\small Global feedback connectivity in experiments with the cooperative actuation. Robots, equipped with color sensors are marked by circles. Thin arrow points to global propagation of messages, thick arrow points to feedback messages. Behinds are robots that do not have any connections to the rest of the robots - this cluster is disconnected from the swarm.\label{fig:fig1} }
\end{figure}

Example of such an information transfer can be given by experiments with collective actuation that requires coordination between robots equipped with color sensors~\cite{Zetterstrom06}, as shown in Fig.~\ref{fig:fig1}. All these robots use local IR-based communication. In this experiment, the robots-scout, equipped with the color sensor, found the blue object. It sends a request to the swarm and asks about a support -- it looks for robots with a specific functionality (in this case also equipped with the color sensor). The behavior of robot-scout (and also the swarm) depends on the feedback signals of other robots with color sensors: when there are no such robots available, scout will look further; when at least two other robots give the feedback, the scout will wait them. This experiment is typical for heterogeneous systems and can be extended towards larger or smaller scales and different sensing and actuating capabilities.

The mechanism of the feedback signals and corresponding collective decision making process, which is trivial for "large" robots with global RF-communication, represents a challenge for micro-robots. The robots should know when to stop sending, know the recipience or routing information for continuously changing situation in the swarm. Not only communication, but also limited computational resources make coordination extremely hard in micro-systems. Complex symbolic negotiation and coordination strategies known in different communities~\cite{Weiss99} can hardly be applied for such micro-agents.

\section{Proposed Mechanisms}
\label{sec:mechanisms}

The idea is to involve in the coordination and negotiation process low-complex numerical mechanisms~\cite{Kornienko_OS01}, which do not require global communication channels as well as on-board computational devices. The required local computation can be performed directly by hardware (an analog way, e.g. FPAA or analog ASIC) and can be implemented even on molecular level~\cite{Balzani03}. The desired collective behavior of these systems can be achieved by applying the principle of artificial self-organization.
\begin{figure}[h!]
\centering 
\includegraphics[width=0.5\textwidth]{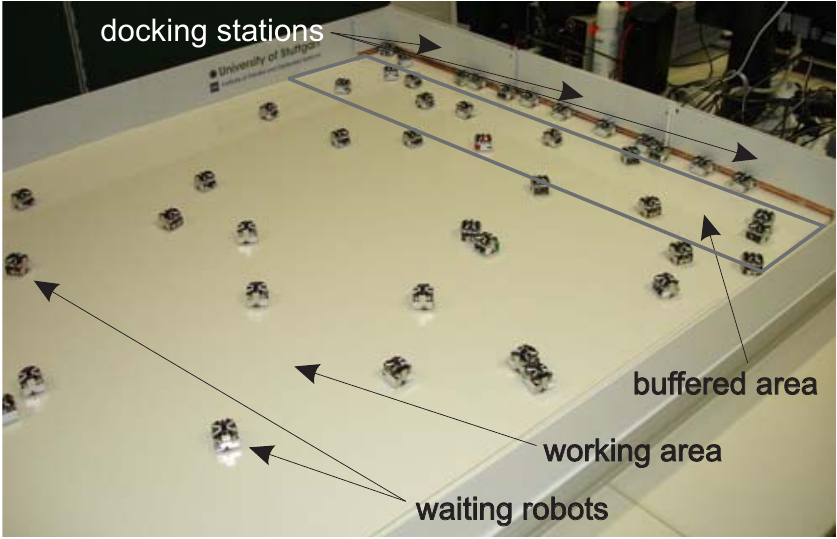}
\caption{\small Collective energy foraging in a swarm of 50 micro-robots "Jasmine". Shown are different roles of robots that are regulated through the low-complex coordination mechanism.\label{fig:fig2} }
\end{figure}

We implemented such mechanisms for the collective energy foraging in the group of 50 micro-robots "Jasmine", shown in Fig.~\ref{fig:fig2}. Each robot has an individual energy homeostasis. When robot is "hungry", it breaks a current activity and starts looking for the docking station. However, when all robots do it in a non-coordinated way, many robots can energetically die due to "bottle necks" at the docking station. The implemented mechanism of low-complex collective decision making provides a coordination mechanism based on the collective power consumption in the swarm. It prevents such situations when many robots simultaneously look for the docking station and maintains the collective energy level of all robots in the optimal way. Here we do not use global information transfer or complex local computations.

\section{Post-Proceeding Comments}

The experiments with collective energy foraging and corresponding bio-inspired and collective decision approaches are published in~\cite{Kornienko_S06b} and~\cite{Kernbach09Nep}. Issues on artificial and adaptive~\cite{kernbach09adaptive} self-organization are explained in~\cite{Kernbach08}, on heterogeneity -- in~\cite{Kernbach09Platform} and~\cite{Kernbach08Permis} and morphology -- in~\cite{Kernbach08_2} and~\cite{Kornienko_S07}. Topics of adaptability in minimalistic swarms are addressed in~\cite{Schlachter08}, \cite{10.1109/ComputationWorld.2009.11}, ~\cite{Kernbach08online} and \cite{kernbach09adaptive}.

\small
\IEEEtriggeratref{20}

\end{document}